\documentclass[conference]{IEEEtran}
\ifCLASSINFOpdf

\else

\fi

\usepackage{amsmath}
\usepackage{xcolor, soul}
\sethlcolor{yellow}
\usepackage{threeparttable}
\usepackage{multirow}
\usepackage{graphicx}
\begin{document}

\title{Accurate Fruit Localisation for Robotic Harvesting using High Resolution LiDAR-Camera Fusion}

\author{\IEEEauthorblockN{Hanwen Kang$^\dagger$, Xing Wang$^\dagger$, Chao Chen$^{*}$}}

\maketitle

\begin{abstract}
Accurate depth-sensing plays a crucial role in securing a high success rate of robotic harvesting in natural orchard environments. Solid-state LiDAR (SSL), a recently introduced LiDAR technique, can perceive high-resolution geometric information of the scenes, which can be potential utilised to receive accurate depth information. Meanwhile, the fusion of the sensory information from LiDAR and camera can significantly enhance the sensing ability of the harvesting robots. This work introduces a LiDAR-camera fusion-based visual sensing and perception strategy to perform accurate fruit localisation for a harvesting robot in the apple orchards. Two SOTA extrinsic calibration methods, target-based and targetless-based, are applied and evaluated to obtain the accurate extrinsic matrix between the LiDAR and camera. With the extrinsic calibration, the point clouds and color images are fused to perform fruit localisation using a one-stage instance segmentation network. Experimental shows that LiDAR-camera achieves better quality on visual sensing in the natural environments. Meanwhile, introducing the LiDAR-camera fusion largely improves the accuracy and robustness of the fruit localisation. Specifically, the standard deviations of fruit localisation by using LiDAR-camera at 0.5 m, 1.2 m, and 1.8 m are 0.245, 0.227, and 0.275 cm respectively. These measurement error is only one one fifth of that from Realsense D455. Lastly, we have attached our visualised point cloud$^{3}$ to demonstrate the highly accurate sensing method.

\end{abstract}


\IEEEpeerreviewmaketitle
\footnotetext[1]{H. Kang, X. Wang, C. Chen are with the Department of Mechanical Engineering, Monash University, Melbourne, Australia.}
\footnotetext[2]{$^\dagger$ Equal Contributions}
\footnotetext[3]{https://drive.google.com/drive/folders/16NV0Bb6N-zlvJC0bFyu-8pl4Gbh9lyOE?usp=sharing}

\section{Introduction}
Robotic harvesting of fruits has shown significant progress in recent development of agricultural industry \cite{vasconez2019agri_robot}. A general procedure for robotic fruit picking requires the localisation then detachment from trees \cite{zhou2021intelligent}. Under the highly complex and unstructured environments in orchards, the accuracy of the fruit localisation is crucial to the performance of the robotic harvesting. Visual perception is the most widely-used method to localize fruits. Significant researches have been conducted to detect and localize the target fruit in 2D color images by means of deep learning \cite{mehta2014vision,si2015location,kang2020fruit}. Additionally, the 3D coordinates is required to conduct the robotic harvesting. In this case. the accurate depth sensing is needed to obtain the information of fruits' location, which is crucial to secure a high success rate of robotic harvesting.

Depth sensing of 3D environments is a fundamental task in robotic vision \cite{liu2020survey}. To achieve a promising performance, various kinds of depth sensors have been developed in past decades, such as monocular camera, stereo depth camera, structural light camera, and Light Detection and Ranging (LiDAR) \cite{horaud2016overview}. The monocular camera can predict the depth value of each pixel, given only a single RGB image as input. Although features like the texture or color of the target can give the information about its depth, the method fails to accurately determine the object's absolute depth \cite{said2012depth}. The stereo depth camera has two or more visual sensors spaced a small distance from each other. The depth of each pixel is computed by stereo disparity through triangulation. The stereo depth camera requires features to associate data, so its accuracy can be severely affected in non-textured or repetitive-textured scenes \cite{luhmann2016sensor}. The structural-light depth camera projects the light with known patterns from an emitter onto the scene. The depth information of the scene is computed by comparing the original pattern with the deformed pattern from the receiver. However, both structural-light depth camera and stereo depth camera have low robustness under complex illumination conditions, which limits their sensing capability in the complex and unstructured agricultural environments \cite{zhong2021survey,maru2020comparison}. LiDAR emits the laser and wavelength, and sweeps them over the scene. The point cloud of the scene is obtained by calculating the time required for each laser beam to get back to the sensor. LiDAR can provide accurate 3D geometric information and be robust to complex illumination conditions. However, the point cloud from a traditional mechanical spinning LiDAR has low resolution, which cannot provide the fine details of a scene \cite{liu2021extrinsic}. Solid-State LiDAR (SSL), a recently introduced LiDAR technique, can provide high-resolution scenes by accumulating point clouds from irregular scanning patterns for a certain period \cite{yuan2021pixel}.
\begin{figure*}[ht]
    \centering
    \includegraphics[width=.95\textwidth]{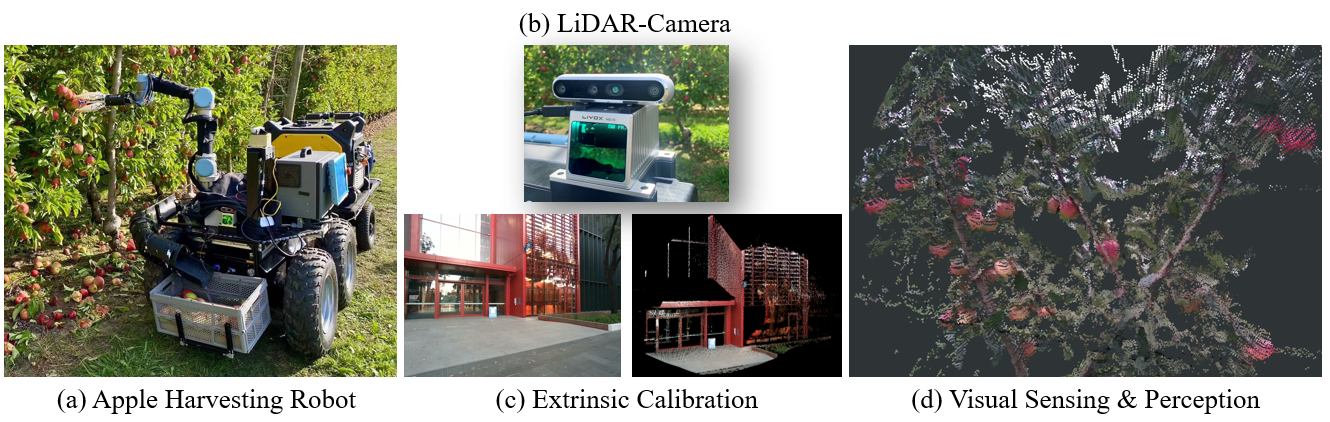}
    \caption{(a) The apple harvesting robot, (b) The attached LiDAR-D455 sensor, (c) Target-less based extrinsic calibration method and its reconstructed scene, (d) Visual sensing of the apple tree in orchard environment}
    \label{fig:system}
\end{figure*}

Even though LiDAR has shown superior performance in perceiving accurate 3D geometries, it cannot capture context and semantic information of the scenes. Fusion of LiDAR and RGB camera is a promising way to achieve better sensing capability for visual perception\cite{zhong2021survey,el2019rgb}. The RGB camera records information by capturing the context of the natural world, while the point cloud directly stores the spatial geometric details of the scenes. However, a preliminary hurdle in LiDAR-camera fusion is to obtain a highly accurate extrinsic calibration between the two sensors. LiDAR-camera calibration can be performed in two steps: feature extraction and extrinsic computation \cite{yuan2021pixel,cui2020acsc}. The first step associates the features from different data expressions, while the second step computes the extrinsic matrix between the sensors. LiDAR-camera fusion is promising and robust for applications that require accurate colorization of dense point clouds, for example, for an automated robotic task within unstructured environments \cite{samal2021task}.

Depth sensing also plays an essential role in visual sensing in conducting agricultural tasks \cite{chene2012use,hassler2019unmanned}. To perceiving the depth information from environments, various depth sensors have been intensively researched, including monocular camera, stereo depth camera, etc. 
Mehta et al. \cite{mehta2014vision} utilized a monocular RGB camera to find the depth distance of citrus relative to the robot base. 
The 2D position is initially obtain using intrinsic parameters of the camera, fruit size, and the pixel coordinates of fruits. The depth information in the camera frame is obtained using triangulation while moving the camera towards the target. The 3D position relative to the robot base was finally calculated using the extrinsic camera parameters. The overall accuracy of the controller was approximately 15 mm, thus making the harvesting of medium or large varieties of citrus fruit possible. Si et al. \cite{si2015location} utilized a stereo camera to detect and locate apples in a natural orchard environment. The apples were detected based on color features, and the detection rate was over 89.5\% with a maximum error of 20 mm at a measuring distance range of 400 to 1500 mm. Our previous works utilized an RGB-D camera, Realsense D435 and D455 to localise apples from the trees \cite{kang2020visual, kang2020real, wang2022geometry}. The depth is received from the aligned depth map from RGB-D camera. In addition, the point cloud from the depth camera was used to estimate each fruit's grasping pose to assist in robotic retrieving. The average accuracy of grasping estimation is 0.61 cm and 4.8$^{\circ}$ in centre and
orientation respectively, at a scanning distance of 0.4m. However, the grasping estimation performance still suffers degradation when the scanning distance increases or under intense sunlight. Another comprehensive comparison of commonly-used depth cameras is conducted and presented in Neupane's work \cite{neupane2021evaluation}.
Depth sensors are also widely applied in other agricultural scenes, such as branch pruning \cite{polic2021compliant}. 

Even though depth sensors have been widely applied for robotic tasks in orchard environment, few works discuss improving visual perception performance by enhancing the sensing capacity of depth sensors. 
With the rapid growth of robotic tasks in the natural orchards, the demand for accurate depth-sensing becomes essential, especially for applications such as robotic harvesting.
To promote the advance of visual perception by enhancing the depth sensing capacity, LiDAR-camera fusion can be potentially studied. This work presents a LiDAR-camera fusion-based depth-sensing strategy to perform accurate fruit localisation in the real orchards environments. The presented method fuses the depth and color data of the environments from a SSL and a camera. It then performs visual perception on fused sensory data to find the precise locations of each fruit. Our method includes two steps: visual sensing and visual perceptions. The visual sensing step utilises a LiDAR-camera fusion to fuse the sensory data from the LiDAR and camera. Before applying data fusion, two SOTA extrinsic calibration methods are applied and evaluated to obtain the accurate extrinsic matrix between the LiDAR and camera. The visual perception step localises the fruits using a one-stage instance segmentation network, which can detect and segment the fruits from the image. Then the depth information from the LiDAR-camera fusion is used to localise the fruits in the task space. To summarise, the following contributions are presented in this paper:
\begin{itemize}
    \item A visual sensing and perception approach that can perform accurate fruit detection and localisation using high-resolution LiDAR-camera fusion method.
    \item A comprehensive study and evaluation on extrinsic calibration strategy of LiDAR and camera.
    \item Demonstration of the presented visual sensing and perception approach on fruit localisation for a robotic harvesting robot in the natural orchard environments.
\end{itemize}

The rest of the paper is organised as follows. The system overview and methodologies of our approach are presented in Section \ref{section: method}. The experiment results and discussion are presented in Section \ref{section: experiments}, followed by the conclusion in Section \ref{section: conclusion}. 

\section{Methodology}\label{section: method}
\subsection{System Setup}
\begin{figure}[ht]
    \centering
    \includegraphics[width=.45\textwidth]{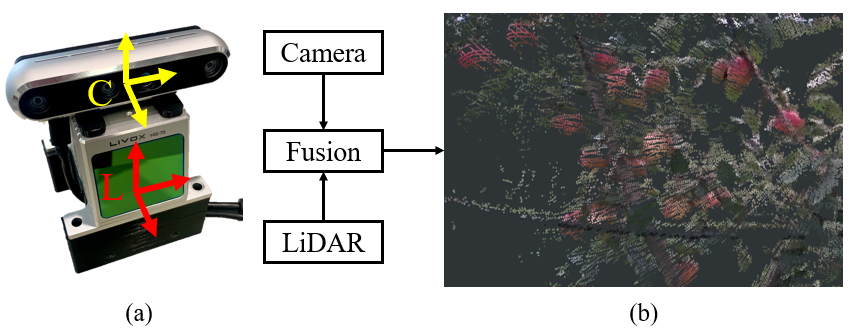}
    \caption{The point cloud from LiDAR and RGB from camera are fused to generate the colored point cloud of the apple tree.}
    \label{fig:lidar-rgb}
\end{figure}

The LiDAR-camera system includes an SSL: Livox Mid-70, and an RGB camera: Intel Realsense-D455, as shown in Figure \ref{fig:system} (b). The LiDAR and camera are installed together with a fixed configuration using 3D printing components. The intrinsic parameters of the RGB camera are evaluated before LiDAR-camera calibration. During the LiDAR-camera calibration, the LiDAR-camera system requires a data acquisition time of 15 seconds for SSL to obtain sufficiently dense points for feature extraction. The LiDAR-camera sensor typically requires less than 2 seconds to acquire sufficient points during visual sensing. Extrinsic calibration plays an essential role in order to accurately matching the depth and color stream from the LiDAR-camera sensor. We comparatively evaluate two SOTA extrinsic calibration approaches: target-based extrinsic calibration and targetless-based extrinsic calibration, which is presented in Section \ref{subsection: calibration}. 
An NVIDIA TX2 is used and connected to the LiDAR and camera to acquire the depth and RGB information simultaneously. We use the LiDAR-camera sensor in a robotic retrieving system to demonstrate the performance of our approach in the in-field agricultural scenarios, which is presented in \ref{subsection: perception}. The pipeline of visual sensing and perception is shown in Figure \ref{fig:lidar-rgb}.

\subsection{Visual Sensing}\label{subsection: fusion}
A pin-hole projection model that project a point $^{C}p_{i} \in R^{3}$ from the LiDAR to a pixel $^{C}\hat{p}_{uv} \in R^{2}$ on color image plane can be computed by using the equation:
\begin{equation}
    ^{C}\hat{p}_{uv} = \pi(K ^{C}p_{i})
\end{equation}
where $\pi( \cdot )$ is the camera distortions model, $K$ is the intrinsic matrix of camera. The  $^{L}p_{i} \in R^{3}$ is the point that acquired by LiDAR. Each point $^{L}p_{i}$ is transformed into camera frame by using the extrinsic matrix $\hat{T}^{C}_{L}$, where $\hat{T}^{C}_{L} = (\hat{R}^{C}_{L},\hat{t}^{C}_{C,L}) \in SE(3)$ (where $\hat{R}^{C}_{L} \in SO(3)$ and $\hat{t}^{C}_{C,L} \in R^{3}$). Therefore, a point $^{L}p_{i}$ that is acquired by LiDAR is projected onto a pixel $^{C}\hat{p}_{uv}$ on RGB image plane through the equation:
\begin{equation}
    ^{C}\hat{p}_{uv} = \pi(K\hat{T}^{C}_{L}\ ^{L}p_{i})
\end{equation}
The intrinsic matrix of the RGB camera is obtained before the LiDAR-camera calibration. The calibrated color and depth image from the LiDAR-camera sensor for visual perception. The obtained colourised point cloud are shown in Figure .

\subsection{LiDAR-camera Calibration}\label{subsection: calibration}
The LiDAR-RGB calibration calculates the extrinsic matrix $\hat{T}^{C}_{L}$ to minimise the reprojection error between image pixels $^{C}p_{uv}$ and the matched projected pixels $^{C}\hat{p}_{uv}$ of the LiDAR points $^{L}p_{i}$, which is formulated as:
\begin{equation}
    min \sum^{n}_{i}||^{C}p_{uv}-^{C}\hat{p}_{uv}||^{2}
    \label{eq: reprojection error}
\end{equation}

Two fundamental problems are solved in LiDAR-camera extrinsic calibration: feature extraction \& data association and extrinsic estimation. The first step extracts features from colour and depth streams. The correspondence between pixels $^{C}p_{uv}$ from the camera and $^{C}\hat{p}_{uv}$ from LiDAR are matched based on the distance or feature similarity. The second step calculates the extrinsic matrix $\hat{T}^{C}_{L}$ between LiDAR and camera by minimising the reprojection error between the matched points and pixels respectively from the LiDAR and camera. Based on the different methods used in the feature extraction \& data association, LiDAR-camera calibration can be divided into target-based calibration and targetless-based calibration. We respectively study and evaluate two calibration methods of target-based and targetless-based methods in this work. The process with better accuracy is used to perform extrinsic calibration of the LiDAR-camera sensor.

\subsubsection{Target-based Method}
Target-based method is commonly used in extrinsic calibration. This method requires a reference target that can be clearly viewed in Field-of-Views (FOVs) of all visual sensors. A recent target-based method, ASAC \cite{cui2020acsc}, is used to calibrate the extrinsic matrix between the LiDAR and camera. 
\begin{figure}[ht]
    \centering
    \includegraphics[width=.47\textwidth]{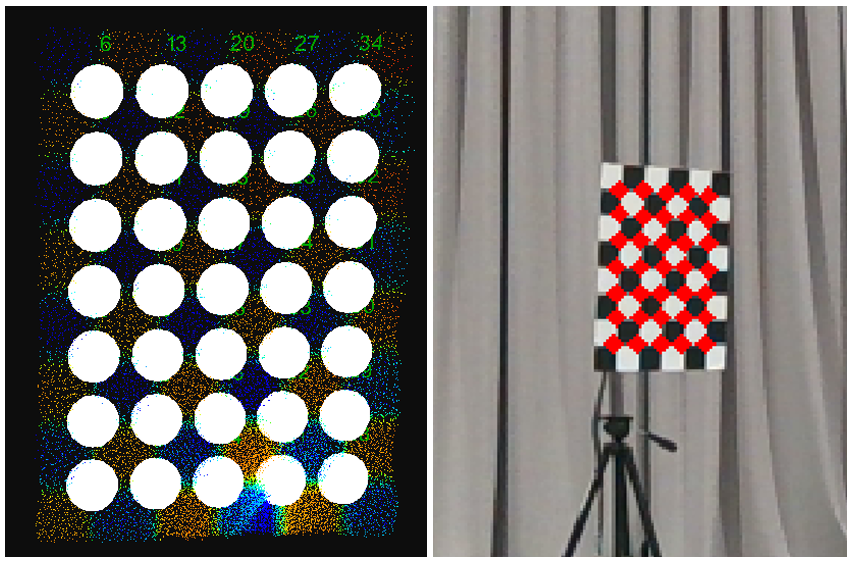}
    \caption{(a) Corner feature extraction (7$\times$5) on the point cloud data, (b) Corner feature extraction on RGB image }
    \label{fig:target}
\end{figure}

It firstly finds the checkerboard corners on point clouds and RGB image, as highlighted in while dots in Figure \ref{fig:target} (a) and red dots in Figure \ref{fig:target} (b), respectively. The checkerboard corners are estimated by optimisation of transformation matrix $\hat{T}^{L}_{M}$ to maximise the similarity between LiDAR points $^{L}P$ and a template checkerboard model points $^{M}S$, which is defined as:
\begin{equation}
    \mathcal{L}(^{M}S,^{L}P|\hat{T}^{L}_{M}) = \sum_{i}^{N_{i}}l_{i}
    \label{eq: target-based cost}
\end{equation}
\begin{equation}
    l_{i} = \sigma(\Bar{p}_{i})\sigma(I_{S}(\Bar{p}_{i}),I_{P}(\Bar{p}_{i}))d(\Bar{p}_{i},G_{i})
\end{equation}
where $N_{i}$ is the number of points in point cloud. $\Bar{p}_{i}$ is the LiDAR point that transformed from the template frame by $\hat{T}^{L}_{M}$.$\sigma(\Bar{p}_{i})$ is the discriminate function to determine whether $\Bar{p}_{i}$ fall in bound of checkerboard area. $\sigma(I_{S}(\Bar{p}_{i}),I_{L}(\Bar{p}_{i}))$ is the discriminate function to identify whether points from template and model have same intensity. $d(\Bar{p}_{i},G_{i})$ measures the distance between $\Bar{p}_{i}$ and the closest corner $G_{i}$ on checkerboard. Equation (\ref{eq: target-based cost}) can be solved by a non-linear optimisation to find correspondence between template and LiDAR. Once the location of each checkerboard corner are determined. The extrinsic matrix between LiDAR and camera can be computed by minimising the reprojection error between matched corners. 

\subsubsection{Targetless-based Method}
\begin{figure}[ht]
    \centering
    \includegraphics[width=.47\textwidth]{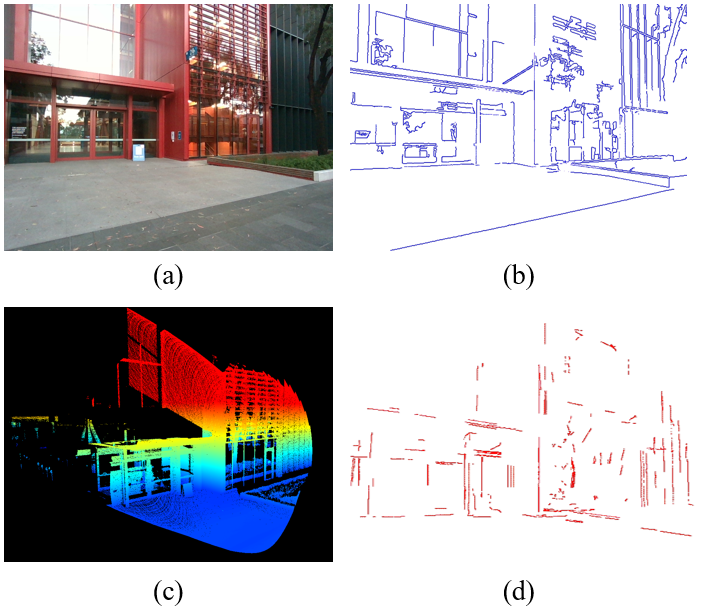}
    \caption{The scene for targetless extrinsic calibration: (a)RGB image, (b)Edges extracted from RGB images, (c)Point cloud in pcd format, (d)edges extracted from the point cloud}
    \label{fig:targetless_f}
\end{figure}

Rather than detecting features from a known geometry, Yuan et al \cite{yuan2021pixel} introduced a tergetless-based method that can calibrate LiDAR and camera by using edge features in general environments. Firstly, edge features are extracted from both image and point cloud. The Canny edge detector is used to extract features from the colour images, as shown in Figure \ref{fig:targetless_f} (b). Then, it down-samples the point cloud into small voxels, and the RANSAC algorithm is used to fit planes and find the depth-continuous intersection lines, as shown in Figure \ref{fig:targetless_f} (d). Only depth-continuous edges from the point cloud are used to avoid significant calibration errors. Correspondences between features are determined by the closest distance by projecting LiDAR points into the image frame using an initial extrinsic matrix. The matched image edges are denoted as $Q_{i}=\{^{I}q_{i};\ i=1,\cdots,k\}$. While the edges points from Lidar are denoted as $P_{i}=\{^{L}p_{i};\ i=1,\cdots,k\}$.

The image edge point $^{I}q_{i}=^{I}\hat{q}_{i}+^{I}w_{i}$, where $^{I}w_{i}\sim\mathcal{N}(0,\sum_{i})$ is the noise associate with point during edge extraction. Let $\hat{d}_{i}$ be the measurement depth and $\sigma_{d_{i}}\sim\mathcal{N}(0,\sum_{d_{i}})$ be the measurement noise. The true depth of $i_{th}$ measurement from LiDAR is given by:
\begin{equation}
    d_{i}=\hat{d}_{i}+\sigma_{d_{i}}
    \label{eq:gt_depth}
\end{equation}
Similarly, let $\hat{w}_{i}$ be the measurement bearing direction and $\sigma(w_{i})\sim\mathcal{N}(0_{2\times1},\sum_{\sigma_{i}})$ be the measurement noise at the tangent plane of $\hat{w}_{i}$. The true bearing direction of measurement $w_{i}$ is obtained by:
\begin{equation}
    w_{i}=e^{([\frac{\hat{w}_{i}}{||\hat{w}_{i}||_{2}}\sigma(w_{i})])}\hat{w}_{i}
    \label{eq:gt_orien}
\end{equation}
where $[\cdot]$ is the skew-symmetric operator. By combining (\ref{eq:gt_depth}) and (\ref{eq:gt_orien}), the edge point of Lidar can be expressed as:
\begin{equation}
\begin{aligned}
    ^{L}p_{i}&=\hat{d}_i\hat{w}_i=(\hat{d}_{i}+\sigma_{d_{i}})e^{([\frac{\hat{w}_{i}}{||\hat{w}_{i}||_{2}}\sigma(w_{i})])}\hat{w}_{i}
\label{eq: lidar point}
\end{aligned}
\end{equation}
Since $e^{[A]}$ is approximately equals $(1+[A])$, the (\ref{eq: lidar point}) can be expand to:
\begin{equation}
    ^{L}p_{i}=\underbrace{\hat{d}_i\hat{w}_i}_{\hat{p}_{i}}+\underbrace{\sigma_{d_i}\hat{w}_i-\hat{d}_i[w_i]\frac{\hat{w}_{i}}{||\hat{w}_{i}||_{2}}\sigma_{w_i}}_{\hat{^{L}{w}_{i}}}
\end{equation}
The projection error of the calibration can be expressed as:
\begin{equation}
    r_i = \pi(KT^{C}_{L}(^{L}\hat{p}_{i}+^{L}w_{i}))-(^{I}\hat{q}_{i}+^{I}w_{i})
    \label{eq: reproject_error}
\end{equation}
The optimisation problem finds the extrinsic matrix $\hat{T}^{C}_{L}$ to minimise reprojection error:
\begin{equation}
    T^{C}_{L}=\mathop{argmin}\limits_{T^{C}_{L}}\sum_{i}^{k}||r_i||_{2}
    \label{eq: opt_1}
\end{equation}
Rather than directly solve the (\ref{eq: opt_1}), authors expand (\ref{eq: reproject_error}) with first order term, leads to
\begin{equation}
\begin{aligned}
    0 &= \pi(KT^{C}_{L}(^{L}\hat{p}_{i}+^{L}w_{i}))-(^{I}\hat{q}_{i}+^{I}w_{i})\\
        &\approx r_i+\mathbf{J}_{T_i}\sigma_T+\mathbf{J}_{w_i}W_i\\
    \label{eq: 1-order term}
\end{aligned}
\end{equation}
where 
\begin{equation}
    W_i={\left[{\begin{array}{c}^{L}w_{i}\\^{I}w_{i}\\\end{array}}\right]}
\end{equation}
Equation (\ref{eq: 1-order term}) implies that:
\begin{equation}
    {r_i}+\mathbf{J}_{T_i}\sigma_{T}=-\mathbf{J}_{T_i}W_i\sim\mathcal{N}(0,\mathbf{J}_{w_i} \sum_{w_i} \mathbf{J}_{w_i}^{T})
\end{equation}
The maximal likelihood with minimum variance extrinsic estimation can be expressed as:
\begin{equation}
\begin{aligned}
    &\mathop{min}\limits_{\sigma_T}\sum_i^k||r_i+\mathbf{J}_{T_i}\sigma_T||_{\sum_{\mathbf{J}_{w_i}W_i}}\\
    &=\sum_i^k(r_i+\mathbf{J}_{T_i})(\mathbf{J}_{w_i} \sum_{w_i} \mathbf{J}_{w_i}^T)^{-1}(r_i+\mathbf{J}_{T_i})
\end{aligned}
\end{equation}
The update policy for $\sigma_T$ is that
\begin{equation}
    \sigma_{T} = (\mathbf{J}_{T}^{T}(\mathbf{J}_{w} \sum_{w} \mathbf{J}_{w}^T)^{-1}\mathbf{J}_{T})^{\dagger}\mathbf{J}_{T}^{T}(\mathbf{J}_{w} \sum_{w} \mathbf{J}_{w}^T)^{-1}r
\end{equation}
$\sigma_T$ is updated iteratively until convergence.

\subsection{Visual Perception} \label{subsection: perception}
\subsubsection{Perception Network}
\begin{figure}[h]
    \centering
    \includegraphics[width=.47\textwidth]{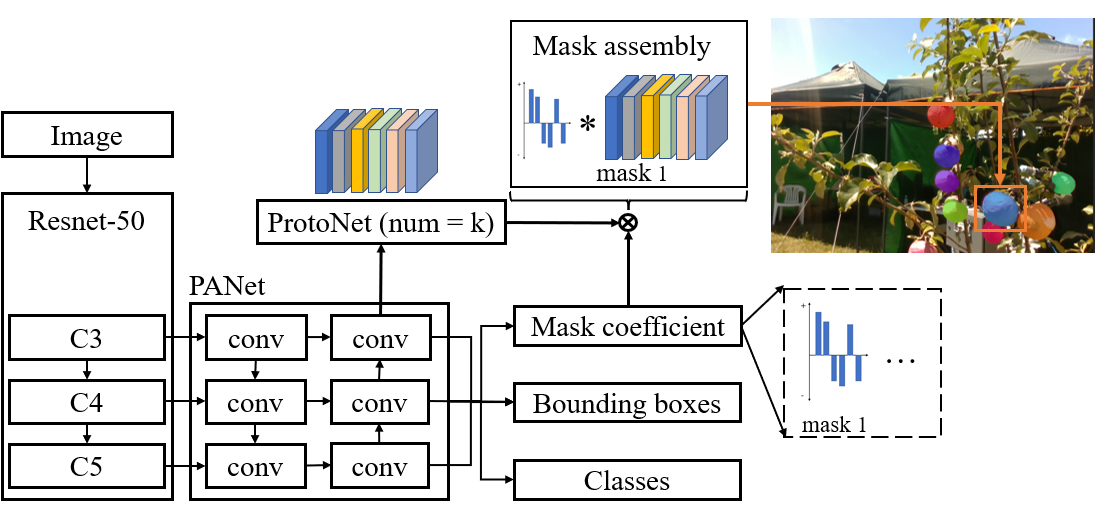}
    \caption{One stage instance segmentation network for apple detection}
    \label{fig:network}
\end{figure}
Fruit recognition and localisation is the core task of the vision system of a robotic system for fruit picking. In this section, we demonstrate the proposed LiDAR-RGB sensor in a robotic fruit picking scenarios. Our method uses an one-stage instance segmentation network architecture, Yolact, to recognise and segment the fruits' mask from the RGB images. Our previous works \cite{wang2022geometry} presented a modified Yolact architecture to perform real-time instance segmentation of the fruits. Compared to the original Yolact, our network architecture can achieve equal or even better accuracy in fruit recognition with better inference efficiency. The network has three sub-networks (as shown in Figure \ref{fig:network}), including a backbone for feature processing, a detection branch for detect fruits from feature maps, and a ProtoNet branch for mask generation. ResNet-50 is used as the backbone of network as it can achieve balance performance on both of accuracy and computational efficiency. A PANet is used in detection branch to fuse feature maps from C3, C4, and C5 of the backbone. Then, each level of output of PANet has a detection head to predict class, Bounding Box (bbox), and ProtoNet coefficients of each object. ProtoNet predicts instance masks of objects follows the principle of semantic segmentation network, which generates a feature map $mask$ of size of $H \times W \times k$, where $H$ and $W$ are height and width of the feature map, $k$ is the number of mask coefficients. The assembly of objects' mask is operated by matrix multiplication and sigmoid activation,
\begin{equation}
    M=\sigma(mask \cdot C^{T})
\end{equation}
where $C$ is the $n \times k$ mask coefficients for $n$ objects. The final instance segmentation mask of each objects are then cropped by using bbox information.

\subsubsection{Network Training}
The cost function to train the network including three parts, which are objects' confidence loss, bounding box localisation loss, and objects' mask coefficient loss. The training cost is formulated as below:
\begin{equation}
\begin{aligned}
    \mathcal{L}=&\lambda_{obj}\sum \mathbf{1}^{obj} L_{obj}+\lambda_{nobj}\sum (1-\mathbf{1}^{obj}) L_{nobj}\\
    &+\lambda_{bbox}\sum \mathbf{1}^{obj} L_{bbox}+\lambda_{mask}\sum \mathbf{1}^{obj} L_{mask}
\end{aligned}                    
\end{equation}
\begin{equation}
    \mathbf{1}^{obj} = 
    \left\{\begin{aligned}
        1 &\quad obj\\
        0 &\quad nobj
       \end{aligned}
       \ \right.
\end{equation}
where $ L_{obj}$ and $ L_{nobj}$ are the objects' confidence loss, $L_{bbox}$ and $L_{mask}$ are respectively object mask coefficient loss and objects' mask coefficient loss. During the training, We set $\lambda_{obj}$ ,$\lambda_{nobj}$, $\lambda_{bbox}$, and $\lambda_{mask}$ as 2.5, 1.0, 1.0, 1.0, respectively. We train the network by using Adam-optimizer for 100 epoches. Since the weights in backbone is pre-trained by using the ImageNet dataset, we frozen the weights of the backbone in the first 70 epoches and train the whole network weights in the last 30 epoches. The learning rate and decay rate are set as 0.001 and 0.9, respectively, in the training.
\begin{figure}[ht]
    \centering
    \includegraphics[width=.49\textwidth]{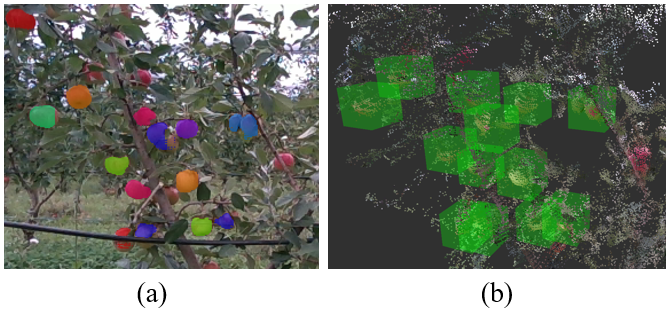}
    \caption{(a) apple instance segmentation on RGB image, (b)Projection of instance masks on point cloud, while green box indicates the detected apples }
    \label{fig:detection_3D}
\end{figure}

\subsubsection{Fruit Localisation}
During the vision perception, the LiDAR-RGB sensor can output the matched color and depth images by using the LiDAR-RGB fusion algorithm which is presented in previous section. From the network inference, we obtained a list of detected fruits with their correspond mask on the RGB image. With the information of the fruits' position and shape on color image, the instance mask of each fruit is projected onto the matched point cloud. Therefore, the point cloud of each fruits can be computed, as shown in Figure \ref{fig:detection_3D}. In another word, the instance mask pixels of a object $O_{i}$ in image are $^{C}Q^{mask}_{i}$. We find the correspond pixels from the matched depth image and compute their position $^{C}P^{mask}_{i}$ in 3D space. Since this work focus on evaluation of fruit localisation accuracy, we use the centre of the point cloud of each fruits as the location of the fruits. In real picking cases, further grasping pose estimation need to be considered to ensure that robot can successfully access and detach the fruits from the tree. A Z-score algorithm is used to reject outliers in $^{C}P^{mask}_{i}$ due to measurement errors, such as depth discontinuous error. The estimated location $^{R}p^{O}_{i}\in R^{3}$ of the fruits then be transformed into the robotic frame by using the equation as below:
\begin{equation}
    ^{R}p^{O}_{i}=T^{R}_{C}\ ^{C}\Bar{P}^{mask}_{i}
\end{equation}
where $T^{R}_{C} \in SE(3)$ is the transformation matrix of camera frame in the robotic frame. 

\subsection{Implementation Details}
Both LiDAR and camera are connected to a Nvidia TX2, which forms the centre control of the proposed vision system. The communication between the LiDAR-RGB sensor and TX2 is achieved by Robotic Operation System (ROS) in Melodic version. The data communication between the TX2 and the LiDAR is achieved by using the Livox-ROS-SDK. The data communication between the TX2 and the Intel Realsense-D455 is achieved by using the Realsense-ROS-SDK. The LiDAR-RGB fusion and calibration are also programmed and running in ROS-melodic. The vision perception network model is programmed by using the Tensorflow-1.15. The training of the network are performed by using a GTX-1080Ti GPU and inference is running on the TX2. 

\section{Experiments and Results}\label{section: experiments}
\subsection{Experimental Setup}
In this study, Livox Mid 70 and Realsense D455 camera are utilized as the LiDAR and RGB module, respectively. They are attached together to maintain relatively fixed position. After the attachment, the RGB camera is calibrated to obtain the intrinsic parameters and distortion coefficients. Next, two extrinsic calibration methods are utilized to find the extrinsic matrix between the Lidar and RGB modules. Lastly, three experiments are conducted to evaluate the accuracy of two extrinsic calibration methods, the accuracy of depth perception by LiDAR-camera system in different indoor scenes, and the fruit detection accuracy in the natural orchard environment, respectively.

\subsection{Camera Intrinsic Calibration}
\begin{figure}[ht]
    \centering
    \includegraphics[width=.45\textwidth]{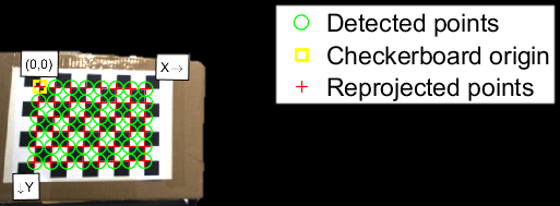}
    \caption{Calibration of RGB module with checkerboard}
    \label{fig:checkerboard}
\end{figure}
The calibration on the RGB camera is firstly required to obtain its intrinsic parameters, including the 3$\times$3 camera intrinsic matrix and the distortion coefficients, which is performed via the MATLAB calibration toolbox. A checkerboard with 10$\times$7 grids is placed on the cardboard, which is then placed within the FoV of RGB module. The image of the checkerboard is then taken for image processing, after which the checkerboard is moved and re-orientated to capture abundant image information. More than 20 images are taken and imported into the camera calibrator toolbox of MATLAB. After checking the corner extraction results, an automatic intrinsic calibration is conducted ending up with results shown in Figure \ref{fig:checkerboard}.
\begin{figure}[ht]
    \centering
    \includegraphics[width=.45\textwidth]{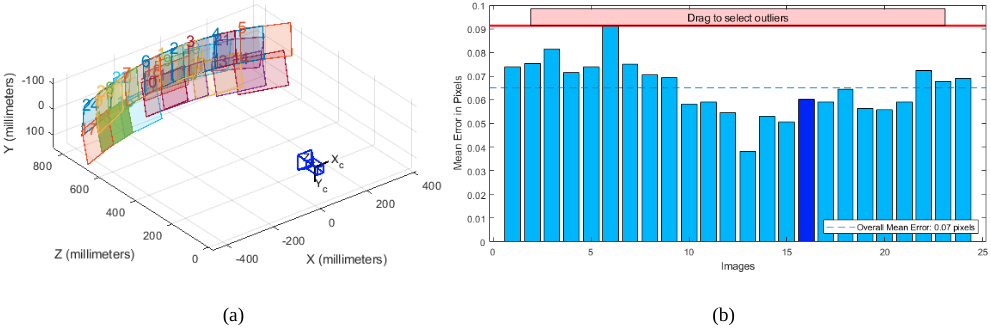}
    \caption{(a) The relative position of RGB module and captured images, (b) The reprojection error of camera intrinsic calibration}
    \label{fig:error}
\end{figure}

The relative positions and orientations between the captured images and the camera are in Figure \ref{fig:error}. Additionally, a reprojection error can be visualized in Figure \ref{fig:error}, from which the maximum reprojection error can be observed to be less than 0.09 pixel. This small reprojection error is critical to guarantee an accurate intrinsic calibration for RGB camera. Finally, the intrinsic matrix and the distortion coefficients can be exported and utilized in the following experiments. 

\subsection{Comparison of LiDAR-RGB Extrinsic Calibration Methods}
This section details the experimental results of the extrinsic calibration. Note that with the relatively fixed position, an initial extrinsic matrix between the LiDAR and camera can be estimated from the CAD file. For further data collection with the integration of LiDAR and camera, it is placed on a tripod and started simultaneously to capture both the RGB image and point cloud data. The collection time for each scene is set to be 15 seconds to allow sufficient numbers of stack frames of the point clouds collected for calibration. Different setups are utilized to conduct the target-based and target-less calibration methods, respectively. 

\begin{figure}[ht]
    \centering
    \includegraphics[width=.47\textwidth]{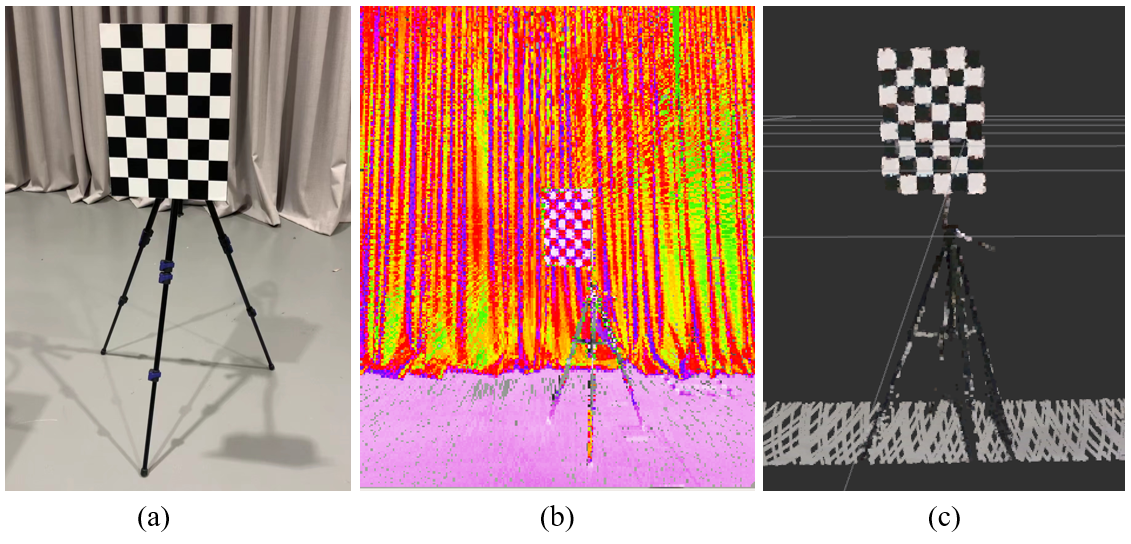}
    \caption{Target-based calibration: (a) RGB image of the checkerboard, (b) Projection of depth information on RGB image, (c) colored point cloud}
    \label{fig:targetbased}
\end{figure}

For the target-based method, a checkerboard is used as the reference target, with a dimension of the inner corners as 7$\times$5 and an inner grid size of 5 cm. Various placements of checkerboard within the common FoV are conducted, including a different range of positions and orientations. To guarantee a successful calibration, some requirements on the selection and placement of the target need to be noticed beforehand. The checkerboard needs to be printed on white paper that is fixed on a non-deformed surface. It should hang in the air with a lower edge parallel to the ground. The final calibration results are visualized in two ways: projecting the point clouds on top of the RGB image Figure \ref{fig:targetbased} (b), and providing the point clouds with colour information Figure \ref{fig:targetbased} (c). The obtained extrinsic matrix is recorded for further accuracy evaluation.

\begin{figure}[h]
    \centering
    \includegraphics[width=.49\textwidth]{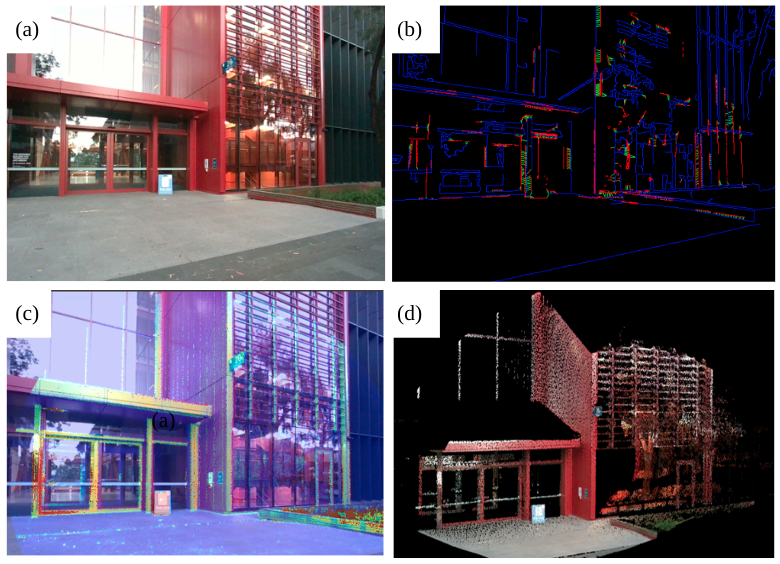}
    \caption{The scene and results for targetless extrinsic calibration with : (a)RGB image, (b)Edges extracted from RGB image (blue) and point cloud (red), (c)Projection of depth information on RGB image, (d)Colourised point cloud with the calibrated extrinsic matrix}
    \label{fig:targetless_visual}
\end{figure}

For the targetless-based method, several scenes with abundant geometrical features are chosen to calibrate the LiDAR-camera. It should be noted that there are requirements for the selection of scenes for the targetless-based method. For instance, scenes with purely cylindrical objects like pillars are not suitable due to the plane fitting requirement of edge extraction. Besides, the extracted edges need to distribute relatively evenly within the FoV to reduce the measurement noise. Lastly, edges should distribute in different directions instead of one like vertical distribution. The initial extrinsic matrix is obtained based on the CAD model and further optimized based on the edges matching. A example of LiDAR-camera extrinsic calibration with origin RGB image, extracted features, depth projection on RGB and colourised point cloud by using targetless-based method are shown in Figure \ref{fig:targetless_visual}.


\begin{table}[ht]
    \centering
    \caption{Comparison of reprojection error}
    \begin{tabular}{c c c}
    \hline
    Method & AVG/pixel & NRE/pixel \\
    \hline
    Target-based & 2.03& 0.78\\
    Target-less &1.91& 0.66\\
    \hline`
    \end{tabular}
    \label{table: comparison of error}
\end{table}
Except for this qualitative evaluation, Normalized Reprojection Error (NRE) is applied to evaluate the accuracy of both target-based and target-less based methods qualitatively. To be specific, we re-project the estimated 3D corners of checkerboard to 2D space with the obtained intrinsic and extrinsic matrix. The corners from RGB images are utilized as the pseudo ground truth to calculate the reprojection error. Additionally, the reprojection error is sensitive to the scanning distance, for example, the error will be smaller than actual value if it is placed too far. So, the error is normalized by multiplying the ratio between scanning distance and maximum distance to eliminate the effect of scanning distance \cite{cui2020acsc}. Note that different extrinsic matrix are utilized from target-based and target-less calibration methods, with the same camera intrinsic matrix. Finally, the results for the NRE are summarised as follows.


\subsection{Evaluation of LiDAR-camera Fusion}
After the calibration as mentioned above, the calculated intrinsic parameters of the camera and extrinsic matrix of the LiDAR-camera are used to fuse the colour and depth streams. This section evaluates and compares the accuracy of depth sensing by using the proposed LiDAR-camera and the Realsense-D455, respectively. Realsense D455 depth camera is a stereo depth camera that has been widely utilized to perform depth-sensing in many robotic applications. Firstly, Table \ref{table: comparison of config} gives the configurations of the LiDAR and Realsense D455.
\begin{table}[ht]
    \centering
    \caption{Comparison of technical specifications of depth sensors used in this study.}
    \begin{threeparttable}
    \begin{tabular}{c c c c}
    \hline
    Sensor model & RS-D455 & Livox-Mid70 \\
    \hline
    Detection range/m  &0.6-6 & 0.2-20\tnote{1} \\
    Depth image (pixel)  &up to 1280x720  & - \\
    Point rate  &- & 100,000 points/s \\
    Range error (0.6-4m)  &$\leq$2\% & - \\
    Range error (0.2-20m)  &- & $\leq$2cm \\
    FoV (H$\times$W)  & 87$^{\circ}\times$ 58$^{\circ}$ & 70.4$^{\circ}$ circular \\
    \hline
    \end{tabular}
    
    \begin{tablenotes}
    \footnotesize
    \item[1]: This can go up to 260m @ 80\% reflectivity. 0.2-20m is to keep error less than 2cm 
    \end{tablenotes}
    \end{threeparttable}
    \label{table: comparison of config}
\end{table}

\begin{figure}[ht]
    \centering
    \includegraphics[width=0.47\textwidth]{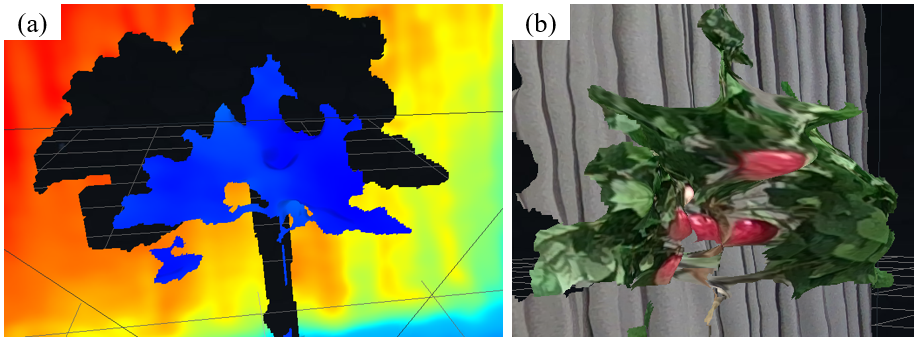}
    \caption{The in-lab tests of fake apple trees by Realsense D455: (a) The recorded depth map, (b) The zoom-in view of colored 3D tree}
    \label{fig: fusion_result_indoor_rs}
\end{figure}

\begin{figure}[ht]
    \centering
    \includegraphics[width=0.47\textwidth]{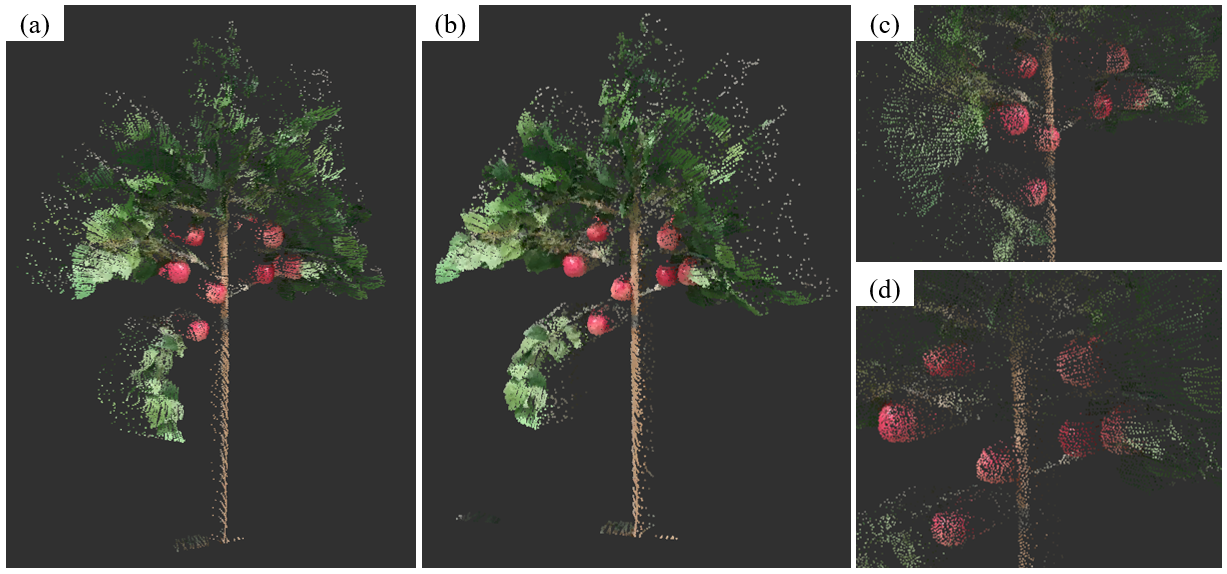}
    \caption{The in-lab tests of fake apple trees by LiDAR-Camera: (a) front view, (b) side view, (c) cropped view, (d) zoom-in view}
    \label{fig: fusion_result_indoor_lc}
\end{figure}

\begin{figure}[ht]
    \centering
    \includegraphics[width=0.47\textwidth]{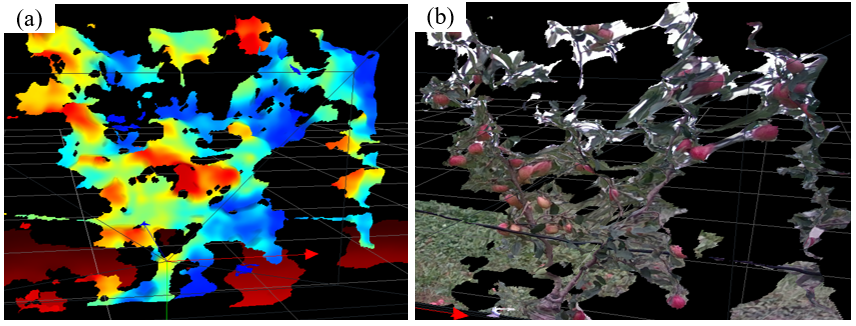}
    \caption{The apple tree detected by Realsense D455 in Fankhauser apple farm (a)depth map, (b)colored point cloud with diffuse lighting}
    \label{fig: fusion_result_outdoor_rs}
\end{figure}

\begin{figure}[ht]
    \centering
    \includegraphics[width=0.47\textwidth]{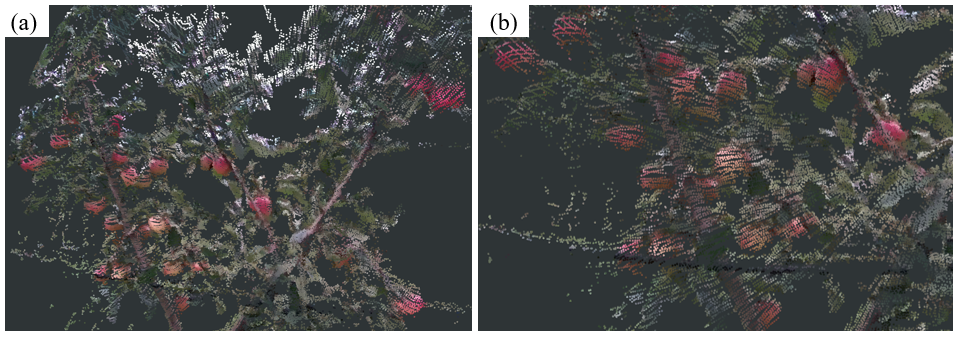}
    \caption{The apple tree detected by Realsense D455 in Fankhauser apple farm: (a) colored point cloud, (b) zoom-in view of colored point cloud}
    \label{fig: fusion_result_outdoor_lc}
\end{figure}

Realsense-D455 has a higher point rate and a much smaller detection range compared to the Livox-Mid70. In the fruit detection and localisation task during robotic harvesting, a detection range of up to 6 meters is sufficient. Regarding the sensing error in the distance, Livox-Mid70 has better accuracy than the Realsense-D455. The depth-sensing performance is qualitatively evaluated by visualising the fusion results of depth and colour streams from the LiDAR and camera. Indoor and outdoor environments are included in this experiment. The fusion results of depth sensing and colour information of a fake apple tree in an indoor scene by using Realsense-D455 and LiDAR-camera are respectively shown in Figure \ref{fig: fusion_result_indoor_rs} and Figure \ref{fig: fusion_result_indoor_lc}. The fusion results of depth sensing and color information in real orchard by using Realsense-D455 and LiDAR-camera are respectively shown in Figure \ref{fig: fusion_result_outdoor_rs} and Figure \ref{fig: fusion_result_outdoor_lc}.

Figures \ref{fig: fusion_result_indoor_rs} and \ref{fig: fusion_result_indoor_lc} show the fusion result of a fake apple tree in the indoor environments at the distance of 1.8 meters. It can be seen that the region of the fake trees can be clearly observed from the point cloud of both Realsense-D455 and the LiDAR-camera. The point cloud from LiDAR-camera has fine geometrical details on the fruits, branches, and leaves. Their shape can be clearly seen in the (c) and (d) of Figure \ref{fig: fusion_result_indoor_lc}. In contrast, the fruits' shape in the point cloud from Realsense-D455 (as shown in Figure \ref{fig: fusion_result_indoor_rs} (b)) is distorted, and the edge between the fruits and other components cannot be clearly distinguished. Figures \ref{fig: fusion_result_outdoor_rs} and \ref{fig: fusion_result_outdoor_lc} show the fusion results in a real orchards at the distance of 1.5-2 meters. At this condition, the quality of the fused point cloud from Realsense-D455 are severely affected. The shape of the apple trees and fruits are severely distorted as shown in Figure \ref{fig: fusion_result_outdoor_rs} (b). Although the quality of the fused point cloud from the LiDAR-camera is also affected, the geometric information of the fruits and surrounding objects are still well preserved, as shown in Figure \ref{fig: fusion_result_outdoor_lc} (b).

\subsection{Demonstration in Fruit Localisation}
\begin{figure}[ht]
    \centering
    \includegraphics[width=.45\textwidth]{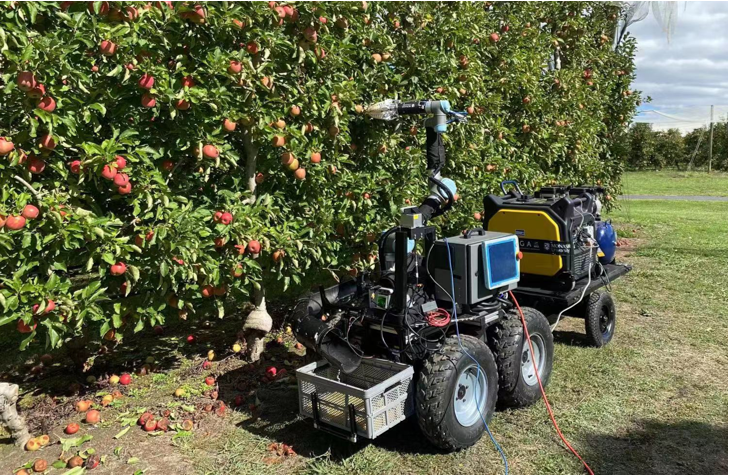}
    \caption{Monash Apple Retrieving System (MARS) operates in Fankhauser apple orchard, Melbourne.}
    \label{fig:mars}
\end{figure}

This section demonstrate the utilisation of the proposed LiDAR-RGB sensor on a robotic fruit picking system, which is developed in our previous works for the autonomous robotic harvesting of apple in orchards, as shown in Figure \ref{fig:mars}.
\begin{figure*}[ht]
    \centering
    \includegraphics[width=\textwidth]{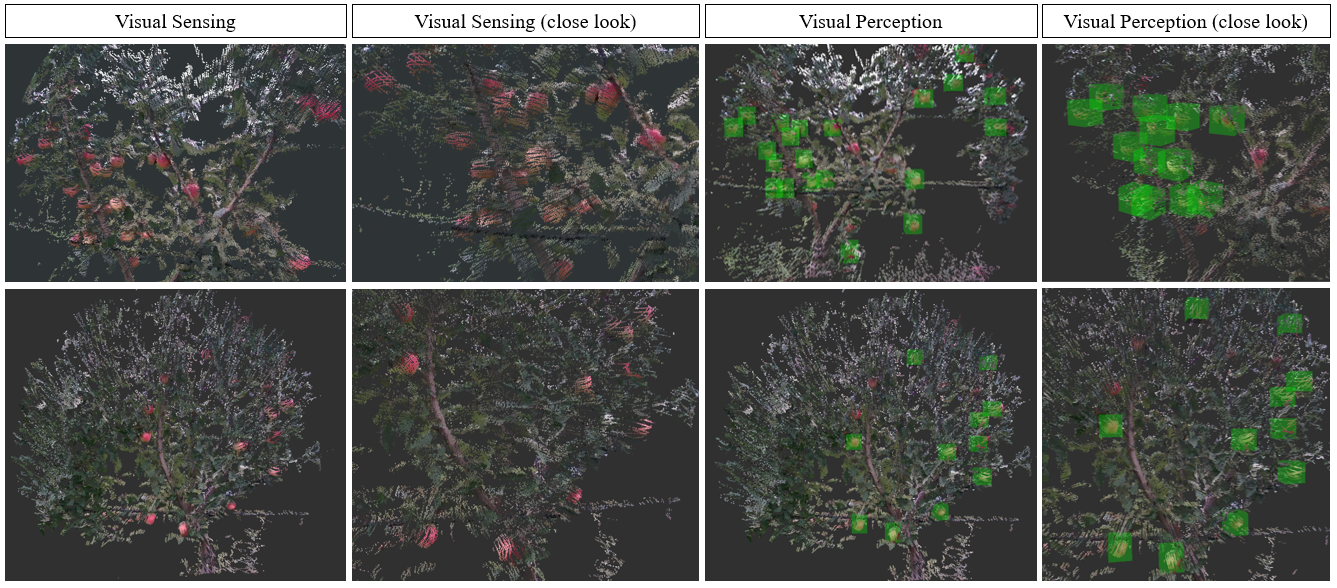}
    \includegraphics[width=0.995\textwidth]{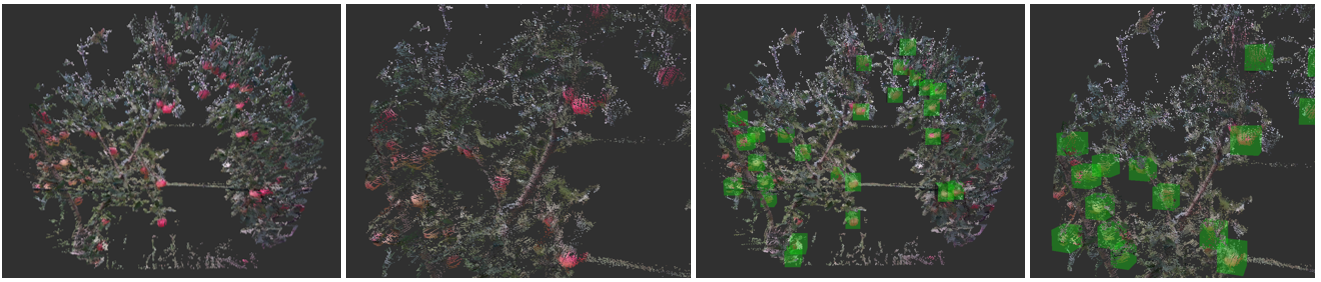}
    \caption{Fankhauser apple orchard: (a)colourised point cloud of apple trees from visual sensing, (b) Closer view of the visual sensing results, (c)Visual perception to detect the apples and project to colourised point cloud, (d)Closer view of the apple detection and projection}
    \label{fig:visual}
\end{figure*}
LiDAR-RGB sensor is used as the "eye" of the system to perceiving the environments. Vision perception algorithms are used to detect and localise the apples from the input sensory data, which forms a critical step before manipulator picking the apple. By using the methods that is described in Section \ref{subsection: perception}, a measurement model for estimation of the fruits' location can be expressed as:  
\begin{equation}
    y_k = x_k+n_k, \ k \ = 0\cdots K 
\end{equation}
where $n_k$ is the measurement noise of the fruit location, which can be expressed as a zero-mean Gaussian distribution, as below:
\begin{equation}
    n_k\in R^3 \sim \mathcal{N}(0,R_k)
\end{equation}
In this experiment, we use $det(R_k)$ to represent the uncertainty within the measurement of the fruits' location. A large value of uncertainty means large error are presented within the measurements. In contract, measurement with higher accuracy will have smaller value on uncertainty. This section evaluates the accuracy of fruits' localisation by respectively using the proposed LiDAR-RGB and Realsense-D455 in natural orchard environments: Fankhauser Apple Farm located in Melbourne, Australia. To eliminate the variables due to other factors such as ambient light, we collect data from the scenes by using the LiDAR-RGB and Realsense-D455 with the same view-angles.

\begin{table}[ht]
    \centering
    \caption{Comparison of depth sensing accuracy of fused sensor and RS D455}
    \begin{tabular}{c | c | c }
    \hline
    \multirow{2}{*}{Sensing Distance}  & \multicolumn{2}{c}{Standard Deviation (cm)} \\
    
     & LiDAR-RGB & RS-D455\\
    
    \hline
    0.5 m & 0.245 & 1.006\\
    
    0.8 m & 0.236 & 1.051\\
    
    1.2 m & 0.227 & 1.175\\
    
    1.5 m & 0.204 & 1.238\\
    
    1.8 m & 0.275  &1.360\\
    \hline
    \end{tabular}
    \label{table: SD}
\end{table}

\begin{figure*}[ht]
    \centering
    \includegraphics[width=1\textwidth]{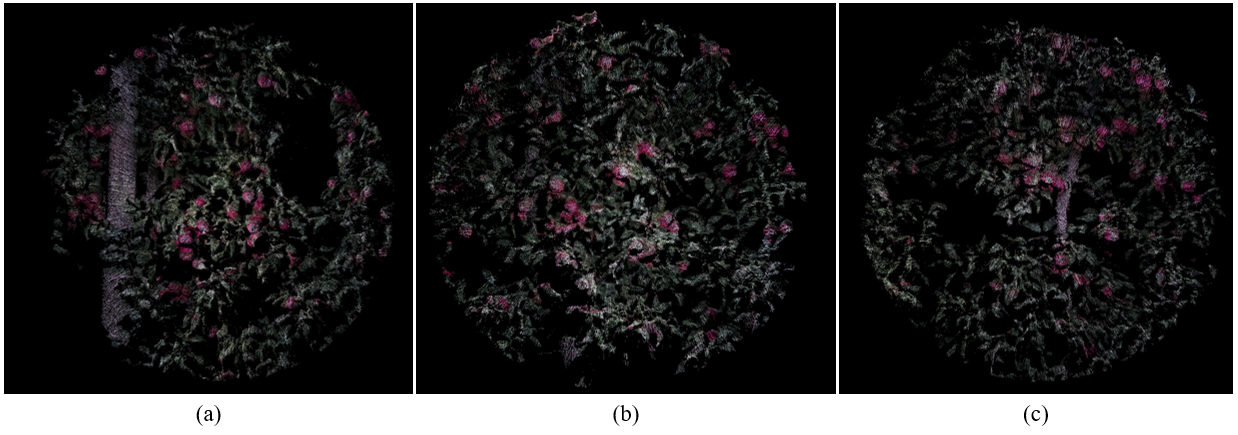}
    \caption{Colourised point clouds of apple trees in Tatura Smartfarm, Australia}
    \label{fig:colored point clouds}
\end{figure*}

The two types of vision sensors are placed at about 0.5m, 1.2m, and 1.8m in front of the tree trunk to evaluate the accuracy of the measurements with the increasing of the sensing distance. The standard deviation (SD) of the uncertainty on the fruit's location is listed in Table \ref{table: SD}. It can be found that the LiDAR-RGB sensor achieves much smaller error in the measurement of the fruit localisation, which is one-fifth compared with the error of Realsense camera at a maximum sensing distance of 1.8 m. Additionally, the SD is similar for the LiDAR-RGB measurement within the proposed sensing range. On the contrary, the SD for Realsense camera's measurement increases with respect to the scanning distance. This is because the scanning accuracy of D455 is proportional to the scanning distance in the depth direction and its accuracy on 2D pixel frame is also coupled with the depth distance. The visualisations of the fruit recognition and localisation at around 1.5 m are shown in Figures \ref{fig:visual} and \ref{fig:colored point clouds}.


\subsection{Discussion}
The experimental results clearly show that LiDAR-camera can perceive accurate depth and context information of the scene compared to the conventional stereo depth camera, such as Realsense-D455. The key to achieving successful visual sensing of the scene requires association between depth and context data from the LiDAR and camera correctly. Therefore, a good estimation of the extrinsic matrix between sensors is essential. Two SOTA automatic LiDAR-camera extrinsic calibration methods are evaluated in this study. Firstly, both work extracts and match the features from the scenes and then calculate the extrinsic matrix by minimising the reprojection error between the corresponding features. From the experimental results, it can be seen that both methods achieve similar accuracy on extrinsic calibration. Comparatively, targetless-based calibration shows better convenience since it does not requires a specific target. Meanwhile, targetless-based calibration can extract features that are evenly distributed in the scenes, which does not require various placements of the reference target within the FoV.

From the comparison of the colourised point cloud from the LiDAR-camera and Realsense-D455, it can be obviously seen that LiDAR-camera can obtain a point cloud with better quality in depth sensing. The geometries of the objects, such as the branches and fruits, within the scenes can be clearly distinguished from the point cloud sensed by the LiDAR-camera. In comparison, most of these geometries are severely distorted in point cloud from the Realsense-D455. The complex geometries and various ambient illumination lead to a significant challenge in perceiving the accurate structure of scenes in an outdoor natural orchard environment. Under this condition, LiDAR-camera shows much better accuracy and robustness than the stereo depth camera. We also notice that the point cloud at the centre region of the LiDAR-camera has better accuracy than the point cloud at the side. This drawback is due to bleeding points caused by laser beam divergence angle. A potential solution to fix this drawback can be achieved by applying multiple LiDAR and only using the point cloud of the centre region from each LiDAR to perform depth sensing.

Finally, the LiDAR-camera demonstrates superior accuracy in fruit localisation for a harvesting robot using a deep-learning detector. The accuracy of the fruit localisation is crucial to secure the success rate and efficiency of the robotic harvesting in the orchard. Meanwhile, a better accuracy on depth sensing can also improve the robotic harvesting strategies by: (1) a better understanding of the geometries of the task space in the real environment, and (2) a more accurate estimation of the proper grasping estimation to avoid collision between robot and trees, which will be concluded in our future works.

\section{Conclusion} \label{section: conclusion}
This study presented a LiDAR-camera fusion-based visual sensing and perception strategy to perform accurate fruit localisation for a harvesting robot in real orchards environments. Two SOTA extrinsic calibration methods, respectively target-based method and targetless-based method, are comprehensively evaluated to calculate the accurate extrinsic matrix between the LiDAR and camera. After calibration, the point clouds and images are fused to perform fruit localisation using a one-stage instance segmentation network. The experimental results show that LiDAR-camera can perform precise and robust depth-sensing in the orchard environments. Meanwhile, introduce of LiDAR-camera largely improve the accuracy and robustness of the fruit localisation. The standard deviations of fruit localisation by using LiDAR-camera at 0.5m, 1.2m, and 1.8m are respectively 0.245, 0.227, and 0.275 m, which significantly outperforms the results by using the Realsense-D455. Future work will focus on further improving the LiDAR-camera extrinsic calibration accuracy. Moreover, the point cloud close to the centre of the LiDAR is more accurate, which indicates that the visual sensing can be enhanced by fusing LiDARs from multiple views. 

\section*{Acknowledgement}
We gratefully acknowledge the financial support from Australian Research Council (ARC ITRH IH150100006).

\bibliographystyle{IEEEtran}
\bibliography{root}

\end{document}